\documentclass[%
 reprint,
 amsmath,amssymb,
 aps,
]{revtex4-1}

\usepackage{graphicx}
\usepackage{dcolumn}
\usepackage{bm}
\nocite{*}

\begin{document}

\title{CogSciK: Clustering for Cognitive Science Motivated Decision Making}
\author{Dr. W. A. Rivera, Director}
\author{James C. Wu, Lead Developer}%
\affiliation{
 Laboratory for Unconventional Conflict Analysis \& Simulation (LUCAS)\\
 Social Science Research Institute\\
 Duke University 
}
\date{\today}
\begin{abstract}
Computational models of decisionmaking must contend with the variance of context and any number of possible decisions that a defined strategic actor can make at a given time. Relying on cognitive science theory, the authors have created an algorithm that captures the orientation of the actor towards an object and arrays the possible decisions available to that actor based on their given intersubjective orientation. This algorithm, like a traditional K-means clustering algorithm, relies on a core-periphery structure that gives the likelihood of moves as those closest to the cluster's centroid. The result is an algorithm that enables unsupervised classification of an array of decision points belonging to an actor's present state and deeply rooted in cognitive science theory.
\end{abstract}

\maketitle

\section{\label{sec:level1}Decision Making Problem}
\subsection{\label{sec:level2}Cognitive Science Paradigm}
Cognitive, Behavioral, and Social Sciences indicate an actor-move hierarchy when performing decision making analysis and simulations. Agents, or entities, in our simulation can be defined with two tiers: actor and strategic actor. The entity "actor" is the simplest actor in the simulation; it has resources and infrastructure but it does not spend or acquire them because it does not make moves relevant to the simulation problem. The strategic actor possesses resources and infrastructure and a wide option of moves with which to acquire or trade away their resources and infrastructure. Throughout the simulation the strategic actor will make moves at each time step based on their present strategic position and desired end-state. 

Moves are defined as a transfer of resources and infrastructure from strategic actors to other actors and entities in the simulation. This simulation utilizes a bank of 374 moves extracted from the Conflict and Meditation Event Observations (CAMEO) codebook and modified by LUCAS. CAMEO moves present a detailed dictionary of move types available to actors. These types may vary between peaceful and aggressive, reckless and timid, short term and long term. Combined they create a robust set of options for strategic actors to pursue strategies through a sequence of moves.

The key decision making problem lies in determining simulation approaches for selecting the strategic actor's next move based on their previous move and their current state. That is, each move is a discrete choice, but must be consistent with goals and consider context. This is a multi-dimensional problem and requires the moves and actor's characteristics to be strictly defined.
\subsection{\label{sec:level2}Intersubjective Orientations}
Intersubjective Orientations (IO) provide a concrete and comprehensive yet flexible way for representing actors in context. The IO values of  Warmth, Affinity, Legitimacy, Dominance, and Competence are taken from Cognitive Science work done by Fiske and Tavares et al., where the value of these variables determine the categorization of agents, events, and other objects in the cognitive space of actors \cite{f1}\cite{t1}. G\"{a}rdenfors work defines these IO values as constituting a decision making space \cite{g2}. Where an object "lands" in the IO space dictates not only the strategic actor's orientation to that object, but what move that strategic actor is likely to take. For instance, an actor with a particular set of IO values towards an object will tend to perform moves with similar IO representations. More specifically if the given IO value indicates high positive values in warmth, for example, hostile attacks such as "attack" are contraindicated. That each move can be uniquely defined by a set of IO values, but also generally interpreted in a group of moves with similar degrees of IO values, provides efficacy and flexibility to the simulation.
\section{\label{sec:level1}Computational Framework}
\subsection{\label{sec:level2}Mapping Actors and Moves}
In order to effectively utilize the IO values in the decision making simulation, they must first be mapped to a range of floating point values. Based on the nature of their interdependence, the IO values are all subject to the same range of possible values. To model the range of emotions associated with these values both negative and positive floating point values are assigned in a scale range of $[-1.0, 1.0]$. Additionally, it is useful to categorize these floating point values into specific classifications to more readily define IO characteristics of actors and moves. This requires another mapping between moves within 5 defined categories $\{A, B, C, D, E\}$, where each category sets specific upper and lower bounds for a move: $A = [-1.0, -0.6], B = [-0.6, -0.2], C = [-0.2, 0.2], D = [0.2, 0.6], E = [0.6, 1.0]$. A move with a Warmth classification of $A$ will have a specific IO value randomly sampled from a uniform distribution bounded by the upper and lower bounds of $A$'s values.

\bigbreak

This random sampling from IO classifications provides clear advantages. First, the random sampling aligns with the problem space in that it ensures no two moves are exactly the same. The overall mapping of IO moves from their IO classifications; e.g. $\{A, C, B, C, E\}$ will yield five randomly sampled floating point values each with eight decimal places of specificity. This procedure ensures that even moves with identical IO classifications will not be identically defined by their IO values, which adds a degree of robustness to the simulation. While it may be the case two moves or two actors are so similar that they have the same IO classification, no two moves and no two actors can be the same, which mimics the complexity of real life decision making. That no moves are identical also yields a simpler computational problem when it comes to defining the rest of an actor or moves' attributes. In essence, an algorithm recognizing the remaining attributes of a move or actor equates to recognizing that move or actor's IO values because only that actor or move can have that particular set of uniquely defined IO values.

\bigbreak

Thus while every move and actor may be rapidly categorized by their IO classification and these classifications can be used to define characteristics between moves, each may also be non-trivially compared to another based on the similarity of their IO values.
\subsection{\label{sec:level2}Similarity Measures}
After mapping the IO's to classifications and values, each actor and move can be represented on a five dimensional space (one dimension for each IO) as a vector of their IO values. Similarly, the distance between actors and moves represents the similarity between the moves or, in the case of actors, the propensity for an actor to make that move based on their IO values. This framework for similarity measures suggests the need for a defined computational method that calculates the distance between each point on the five dimensional IO-space. In developing this computational method, two requirements must be heeded: 

\bigbreak

First, the method must apply a uniform kernel function across all move pairings to evaluate similarities. In both statistics (kernel density estimation or kernel smoothing) and machine learning (kernel methods) literature, a kernel is used as a measure of similarity. In particular, the kernel function $k(x,.)$ defines the distribution of similarities of points around a given point $x$. $k(x,y)$ denotes the similarity of point $x$ with another given point $y$ \cite{c1}. Applying a kernel function requires maintaining consistent evaluations within each individual IO pairing; e.g.: Warmth of one move compared to Warmth of another. This does not restrict the comparisons between the different values; e.g. the method may implement a different comparison function between Warmth values and Legitimacy values. Furthermore the kernel function combines the outputs of these individual comparisons to create a single distance measure that represents the similarity between two points in the IO space. The combination function remains flexible depending upon the individual weighting of IOs; Warmth may be considered as twice as important as Legitimacy, or the similarity measure may simply be a weighted average of the output of all the IO comparison functions. 

\bigbreak

Second, the method must remain computationally inexpensive. Depending on the design of the decision making simulation, the similarity measure will need to be calculated for every move in the system. This not only includes distances between the moves and an actor or multiple actors, represented by their IO vectors, but also the distances between two moves, represented by their IO vectors. 

\bigbreak

Keeping these restrictions in mind, the optimal kernel function for this problem is Euclidean Distance, which is derived from the Euclidean (L2) norm. Euclidean Distance assumes equal weights on each IO when it performs a squareroot on the sum of squared differences between the moves' IO values. The original IO mappings enable this approach because the values are restricted to the same range and scale. Furthermore, this method employs the same distance comparison, the L2 norm, between IOs, which enforces consistency between IO evaluations, and it is a relatively inexpensive computation to repeat many times in the simulation \cite{b1}. 
\section{\label{sec:level1}Traditional Modeling Techniques}
\subsection{\label{sec:level2}K-Means Clustering}
With the selection of Euclidean Distance to evaluate the similarity measure between moves, the next step requires choosing an algorithm to select the optimal move for an actor in a decision making scenario. The decision space currently lacks a defined response for an algorithm to be trained to select, so the only option is an unsupervised approach. K-means clustering is a popular unsupervised clustering algorithm, and it frequently clusters points based on Euclidean distances by creating clusters around initial randomly assigned $k$ centroids \cite{b1}. The algorithm is simple to implement and provides a guaranteed convergence, so it does not delay the decision making scenario. 

\subsection{\label{sec:level2}Application to the Problem Space}
K-means clustering is certainly useful for investigating the decision making space. Creating clusters of moves potentially reveals characteristics about the moves in each cluster that were previously unknown to the simulation. It also provides a way to validate the remaining attributes of the moves based on their IO values and validate whether a moves' IO values are properly categorized. For instance, it seems reasonable that moves that gravitate towards a violent or aggressive type classification will have similar IO values. This will result in the algorithms placing them in the same cluster or nearby clusters, so that each cluster will have an overarching type based on the attributes of the moves within. The centroids will provide the best representation of the overall type and attributes of a cluster. This results in very useful analysis of the present types assigned to moves and may even help detect an error with the random sampling technique in the IO mapping step if a particular move belongs to an unexpected cluster.

\bigbreak

G\"{a}rdenfors extends the cognitive science theories with his representation of IO values as quality dimensions in particular conceptual spaces. “A conceptual space consists of a number of quality dimensions...It is assumed that each of the quality dimensions is endowed with certain topological or metric structures” \cite{g1}. Specifically, the way a move is perceived depends on its particular orientation. Each subject, or actor in this case, perceives orientations in a different way (recall the methodology regarding randomizing IO values). These variations can be due to personal factors regarding the actor or the dynamic space of the decision making space itself, which is comprised of not just actors, but also resources and infrastructure the actors rely upon. Thus constructing a particular move requires a spatial relationship between dimensions that can be computed mathematically. Manipulating dimensions can create a particular IO vector that fits an ideal type of move or actor that may not yet exist empirically in the simulation. The K-means algorithm is built with a core periphery structure such that each generated cluster in the space has a centroid representing primary importance to a strategic actor given by their strategy and then other elements are grouped around the centroid by proximity and necessity to achieve the actor's strategic goals\cite{g2}.

\bigbreak

However, a traditional K means algorithm clustering approach fails to account for the impact of a single actor's IO values. In the present unsupervised framework, the actor may be included into the dataset of moves and treated as a move by the clustering algorithm. While computationally there are no flaws to this approach, it essentially assigns the same importance to an actor as a move, which makes little sense in a decision making simulation where the actor is responsible for selecting moves. Moreover, an actor possesses very different attributes from moves, so the useful techniques associated to the core periphery structure of G\"{a}rdenfors' reasoning that unlocks clustering based analysis of the moveset does not apply to the actor. More likely, the actor's IO values will be drowned in a sea of moves' IO values, diminishing its importance and undermining the cognitive science paradigm for the decision making scenario.
\section{\label{sec:level1}CogSciK}
\subsection{\label{sec:level2}Algorithm and Problem Space Application}
The CogSciK algorithm incorporates many elements of the K-means clustering algorithm and even utilizes a Euclidean Distance function to measure similarity, but it differs in that it takes in a centroid as an initial input along with the training set of moves. This single centroid represents the actor's IO vector and directs the algorithm to provide more emphasis on the actor, which makes the algorithm much more suitable for decision making simulations compared to an algorithm such as K-means, which randomly assigns moves as centroids for the clusters. This allows the algorithm to essentially select a set of moves the actor may pursue in the next steps of the decision making simulation based on the IO values of the actor and moves available.

\bigbreak

The CogSciK algorithm is defined as follows:
\begin{enumerate}
\item Initialize the centroid with the algorithm's input, the actor's IO value.
\item Load the moves data set and perform the IO mappings using the random sampling techniques described.
\item Calculate the similarity between the centroid and each move in the dataset using the Euclidean Distance function.
\item Sort the moves by the smallest calculated distance between the moves and the centroid and select $k$ moves. This is the moves cluster with the actor as a centroid.
\item Use other decision making algorithms in the simulation to select a specific move for the actor to perform in the next step, select the nearest move to the actor, or randomly select the move.
\item Initialize a new centroid and cluster size based on the selected move.
\item Repeat the algorithm for $n$ timeticks.
\item Add additional actors by initializing new centroids based on their IO values and running the algorithm with those centroids as inputs.
\end{enumerate}

Apart from its designed focus towards selecting moves based on actor IO's, the CogSciK algorithm provides several additional benefits that optimize its fit for the decision making problem. That the algorithm can sequentially generate clusters based on similarities models a real life scenario with clearly defined time steps in the simulation. The algorithm also handles move selection for multiple strategic actors by allowing multiple centroids to be initialized and move clusters to be built around them. 

\subsection{\label{sec:level2}CogSciK Package}
The CogSciK package is implemented in Python2.7 and published on GitHub and PyPi as an installable python package. The package contains an abstractable implementation of the entire CogSciK algorithm, so that it may be applied to additional decision making problems. The GitHub tutorial includes a custom Move object and Cluster object that contain the IO and attributes defined in this problem space. An example of the moves cluster generated by the algorithm for an actor with initial IO values $[A, B, C, D, E]$ is below:
\begin{verbatim}

Cluster Primary Type: Reject
	Cluster Size: 10
	Centroid IO: [-0.64693745 -0.44900883 
	              -0.07025932  0.42156327  
	               0.68451422] 

Move: Rally opposition against
Move Type: Disapprove 

Move: Accuse of espionage, treason
Move Type: Disapprove 

Move: Refuse to build energy infrastructure
Move Type: Refuse to build infrastructure 

Move: Conduct suicide, car,
      or other non-military bombing
Move Type: Assault 

Move: Reject economic cooperation
Move Type: Reject 

Move: Appeal to yield
Move Type: Appeal 

Move: Increase censorship
Move Type: Control information 

Move: Make a pessimistic comment
Move Type: Make a public statement 

Move: Reject request for military aid
Move Type: Reject 

Move: Reject request for economic aid
Move Type: Reject 
\end{verbatim}

The cluster info outputted assigns the cluster's overall type to the most frequently occurring move type, Reject. The remaining move types, such as Appeal and Disapprove are similar types to Reject, which makes intuitive sense because the IOs of those move values should be similar to each other. The Move object can be modified to hold different move attributes and even to include additional IO's in the IO space. Similarly, multiple actors can be initialized and inputted simply as vectors of IO values. 

\begin{verbatim}
GitHub: https://github.com/jamescwu/CogSciK
PyPi: https://pypi.python.org/pypi/cogscik/1.0
\end{verbatim}

\newpage
\onecolumngrid
\section{Bibliography}
\bibliography{cogscik} 

\begin{thebibliography}{1}

\bibitem{f1}
S.~Fiske, ``Warmth and competence: Stereotype content issue for clinicians and
  researchers,'' {\em Canadian Psychology}, no.~53, pp.~14--21, 2012.

\bibitem{t1}
R.~M. Tavares, ``A map for social navigation in the human brain,'' {\em
  Neuron}, no.~87, pp.~231--243, 2015.

\bibitem{g2}
P.~G\"{a}rdenfors, ``Mental representations, conceptual spaces and metaphors,''
  {\em Synthese}, no.~106, pp.~21--47, 1996.

\bibitem{c1}
F.~Camastra and A.~Verri, ``A novel kernel method for clustering,'' {\em IEEE
  Transactions on Pattern Analysis and Machine Intelligence}, no.~27,
  pp.~801--805, 2005.

\bibitem{b1}
M.~D.~J. Bora and D.~A.~K. Gupta, ``Effect of different distance measures on
  the performance of k-means algorithm: An experimental study in matlab,''
  2014.

\bibitem{g1}
P.~G\"{a}rdenfors, {\em Conceptual Spaces: The Geometry of Thought}.
\newblock Lawrence Erlbaum, 2000.

\end{thebibliography}
\bibliographystyle{ieeetr}

\end{document}